%% file: acl_latex.tex
\documentclass[11pt]{article}

\usepackage[]{acl}

\usepackage{times}
\usepackage{latexsym}
\usepackage{graphicx}
\usepackage{booktabs}
\usepackage{adjustbox}
\usepackage{enumitem}
\usepackage{multirow}

\usepackage[T1]{fontenc}
\usepackage[utf8]{inputenc}

\usepackage{microtype}

\usepackage{amsmath}

\usepackage{inconsolata}

\usepackage{tabularx}

\title{On the Multi-turn Instruction Following for Conversational Web Agents}

\author{Yang Deng$^1$\thanks{~~ Equal contribution.} , Xuan Zhang$^{1*}$, Wenxuan Zhang$^2$, Yifei Yuan$^3$, See-Kiong Ng$^1$, Tat-Seng Chua$^1$\\
        $^1$National University of Singapore, $^2$DAMO Academy, Alibaba Group, $^3$University of Copenhagen\\ 
        \texttt{ydeng@nus.edu.sg} ~ \texttt{xuanzhang@u.nus.edu}}

\begin{document}
\maketitle
\begin{abstract}
Web agents powered by Large Language Models (LLMs) have demonstrated remarkable abilities in planning and executing multi-step interactions within complex web-based environments, fulfilling a wide range of web navigation tasks. Despite these advancements, the potential for LLM-powered agents to effectively engage with sequential user instructions in real-world scenarios has not been fully explored. 
In this work, we introduce a new task of Conversational Web Navigation, which necessitates sophisticated interactions that span multiple turns with both the users and the environment, supported by a specially developed dataset named Multi-Turn Mind2Web (MT-Mind2Web).
To tackle the limited context length of LLMs and the context-dependency issue of the conversational tasks, we further propose a novel framework, named self-reflective memory-augmented planning (Self-MAP), which employs memory utilization and self-reflection techniques. 
Extensive experiments are conducted to benchmark the MT-Mind2Web dataset, and validate the effectiveness of the proposed method.\footnote{The dataset and code will be released via \url{https://github.com/magicgh/self-map}.}
\end{abstract}

\section{Introduction}
\begin{figure}[t]
\setlength{\abovecaptionskip}{5pt}   
\setlength{\belowcaptionskip}{0pt}
\centering 
\includegraphics[width=0.48\textwidth]{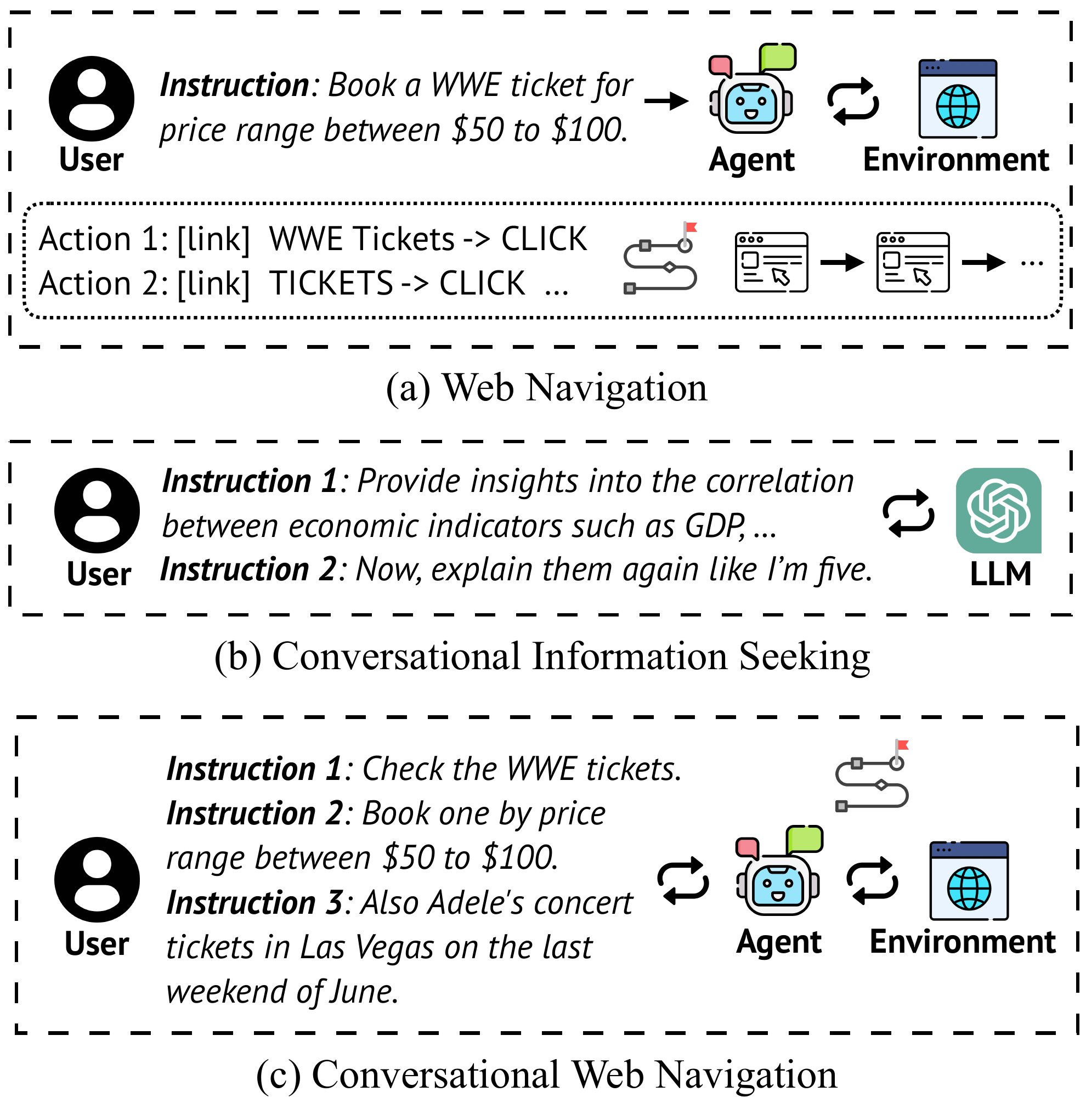}
\caption{Illustrations of different problems.}
\label{fig:example}
\vspace{-0.3cm}
\end{figure}

A longstanding objective in artificial intelligence is to develop AI agents \cite{wooldridge1995intelligent} that can execute complex tasks, thereby minimizing human effort in routine activities. With the advent of Large Language Models (LLMs), LLM-powered agents \cite{ruc-agent-survey,fudan-agent-survey} showcase exceptional planning capabilities in performing multi-turn interactions with diverse environments, which contribute to various real-world problem-solving. As shown in Figure \ref{fig:example}(a), the web agent \cite{mind2web,webarena,webshop} is designed to interpret the states of a webpage and execute a series of actions using keyboard and mouse inputs. Its purpose is to accomplish the tasks defined in natural language, such as booking tickets, through multi-turn interactions with the web-grounded environment.

Despite the proficiency in executing each individual instruction, the capability of interacting with multi-turn user instructions remains under-explored, which is crucial for applying LLM-powered agents onto real-world applications. 
As the example shown in Figure \ref{fig:example}(c), during a conversational web navigation session, users tend to request follow-up or co-referencing instructions without repeating previous information. They may also  provide a succinct or brief instruction, which is similar to other conversation problems. 
Motivated by recent efforts \cite{nips23-mtbench,llm-cis,iclr24-ppdpp} on the investigation of conversational capabilities in the interactions with human users for LLMs, we propose a novel task, named \textbf{Conversational Web Navigation}. It requires the multi-turn interaction capabilities with both users and environment. 
In particular, we introduce a new dataset, named Multi-Turn Mind2Web (MT-Mind2Web).
MT-Mind2Web is constructed by using the single-turn interactions from Mind2Web \cite{mind2web}, an expert-annotated web navigation dataset, as the guidance to construct conversation sessions. 

In other conversational tasks, LLMs can answer conversational questions \cite{nips23-mtbench} by utilizing their inherent knowledge from pretrained data or retrieval techniques to assess external databases (Figure \ref{fig:example}(b)). Compared with these tasks, the conversation history in conversational web navigation contains both the previous user-agent and agent-environment interactions, as the instruction completion relies on the dynamic environment status. Therefore, the history context can be much longer and noisier than that in the traditional conversation problems.

In light of these challenges, we propose a novel framework, named self-reflective memory-augmented planning  (Self-MAP). This framework is designed  to maximize the utility of the limited memory space (\textit{i.e.}, input length limitation) of LLM-powered agents addressing the conversational web navigation problem. 
Specifically, we first construct a memory bank using the conversational interaction history, where each memory snippet stores each interaction step at each conversation turn. To reduce the noise from previous interactions, we propose a multifaceted matching approach to retrieve memory snippets that are semantically relevant and have similar trajectories. 
Furthermore, we design a reflection module to simplify the retrieved memory snippets by filtering out irrelevant information from the environment state. We then refine the retrieved memory snippets by generating reasoning rationales to enrich the memory information.  
Finally, we plan the next action by utilizing the self-reflective memory. 

To sum up, our contributions are as follows:

\begin{itemize}[leftmargin=*]
    \item To study the multi-turn instruction-following capability of web agents, we define the problem of conversational web navigation and introduce a novel dataset, namely MT-Mind2Web. 
    \item We propose a self-reflective memory-augmented planning method (Self-MAP) that combines memory utilization and self-reflection for tackling the underlying challenges in the conversational web navigation task. 
    \item We benchmark the MT-Mind2Web dataset with extensive baselines and provide comprehensive evaluations on different settings. Experimental results also validate the effectiveness of the proposed method. 
\end{itemize}

\section{Related Works}

\paragraph{Web Agents}
Evolving from web agents with simplified environment simulation \cite{icml17-webagent,iclr18-webagent,naacl21-webagent,webshop}, recent studies investigate web navigation problems under more practical and complex settings, including multi-domain \cite{mind2web}, real-time interactions \cite{webarena}, and visual UI understanding \cite{seeact}.  
To handle these advanced web navigation problems, there has been increasing attention on building autonomous web agents powered by LLMs \cite{ruc-agent-survey,fudan-agent-survey}. 
Various prompt-based methods have been proposed to enhance the LLM-powered web agents, such as recursive self-correction prompting \cite{nips23-rci}, code-based prompting \cite{adaplanner}, and trajectory-augmented prompting \cite{iclr24-trajectory}. 
However, prompt-based methods typically fail to compete with fine-tuned methods \cite{webagent-htmlt5,mind2web} in advanced settings, such as Mind2Web. 
In this work, we propose a new task, namely conversational web navigation, which requires multi-turn interaction capabilities with both users and the environment. 

\begin{figure*}[t]
\setlength{\abovecaptionskip}{2pt}   
\setlength{\belowcaptionskip}{2pt}
\centering 
\includegraphics[width=\textwidth]{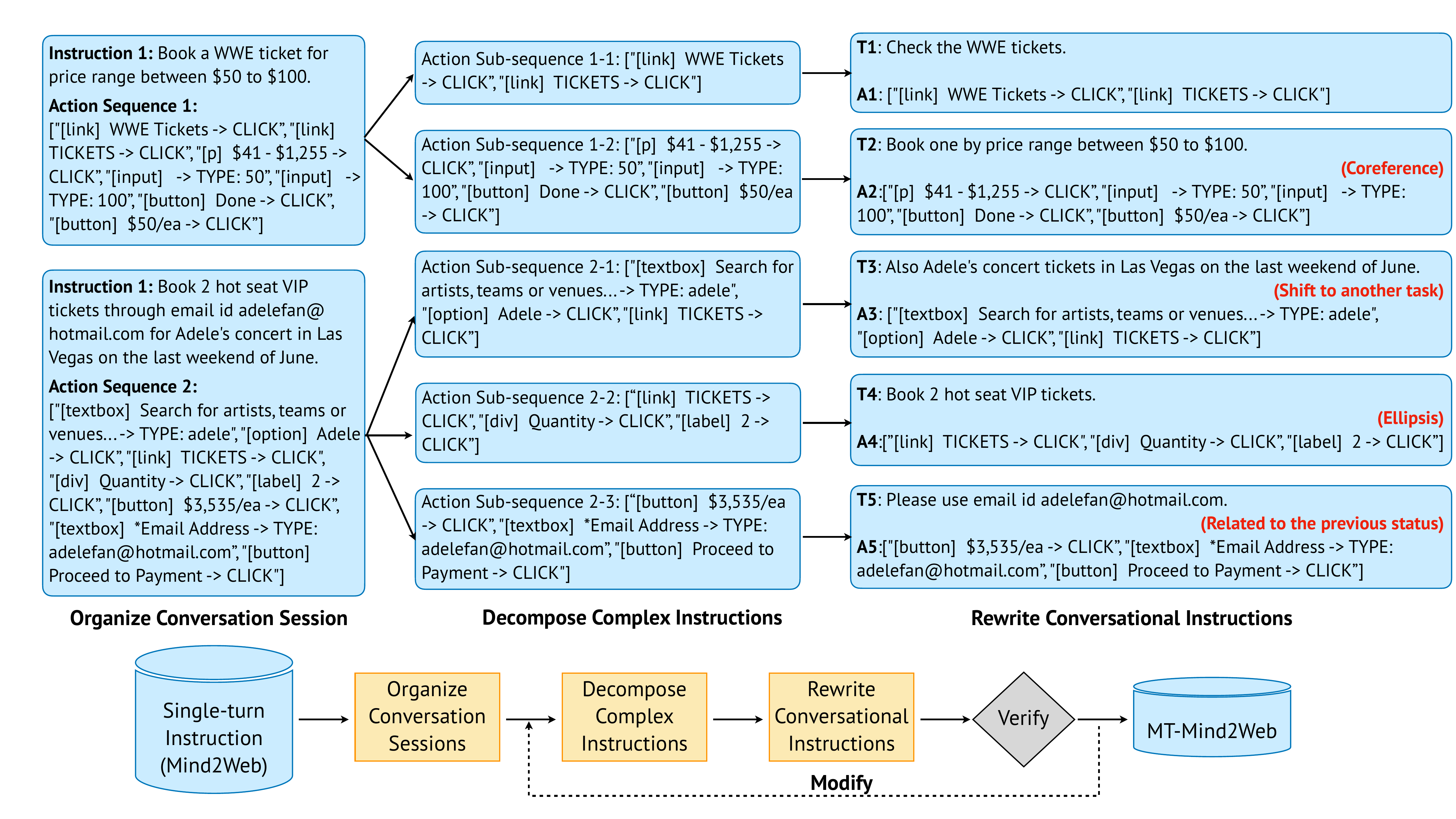}
\caption{Overall pipeline for MT-Mind2Web creation with examples. }
\label{fig:pipeline}
\vspace{-0.3cm}
\end{figure*}

\paragraph{Multi-turn Interactions with Environment}
Interacting with the external environment enables LLM-powered agents to handle challenging tasks \cite{agentbench,ma2024agentboard}.  For example, agents can interact with a code-grounded environment to access databases or perform programming \cite{iclr24-Lemur,metagpt}, game-grounded environment to foster entertainment \cite{alfworld}, web-grounded environment to navigate webpages \cite{mind2web} or perform online shopping \cite{webshop}. 
These works mainly focus on completing a standalone user instruction by planning a sequence of actions to interact with the environment. 
Some latest studies \cite{mint,openagents} investigate the utilization of multi-turn user feedback for solving a given task. 
In real-world applications, users may not always ask for the assistance for only a single task, while follow-up instructions and multi-turn requests are common during a conversation session.

\paragraph{Multi-turn Interactions with Users}
Extensive studies demonstrate the exceptional capabilities of LLMs in seamless multi-turn interactions \cite{nips23-mtbench} with human users for completing various conversational tasks during a conversation session, such as  recommendation \cite{cikm23-llm-crs,recagent}, tutoring \cite{educhat,iclr24-ppdpp}, counseling \cite{chatpal}. 
For instance, MT-Bench \cite{nips23-mtbench} is one of the most popular benchmarks for evaluating the multi-turn instruction-following ability of LLMs. It consists of 80 high-quality multi-turn questions ranging from 8 common instruction-following abilities, such as writing, roleplay, reasoning, etc. 
However, these conversational tasks mainly rely on the inherent knowledge of LLMs or just perform a one-time request from the external environment for each turn, such as conversational information seeking \cite{llm-cis}, without the need to access the dynamic environment for multiple times.

\section{MT-Mind2Web Dataset}

\subsection{Annotation \& Quality Control}
Inspired by the typical construction process of existing conversation datasets, such as HybriDialogue~\cite{hybridialogue} from OTT-QA~\cite{ott-qa}, MMCoQA~\cite{mmconvqa} from MMQA~\cite{mmqa}, and PACIFIC \cite{pacific} from TAT-QA \cite{tat-qa}, we build the MT-Mind2Web dataset from the Mind2Web dataset \cite{mind2web} by using its single-turn interaction data as guidance for constructing conversation sessions. 
In order to reuse the expert-annotated action sequences in Mind2Web for ensuring the system response quality, the conversation construction process mainly focuses on the user instructions. 
In specific, the construction process contains three main steps: 

\paragraph{1) Organize Conversation Sessions}
Given the same context, \textit{i.e.,} the same domain and website in Mind2Web, set up a conversation session with consecutive topics from multiple individual task instructions. Two instructions that share the same entities or intents are regarded as talking about the same topic. As the example in Figure \ref{fig:pipeline}, both the original \textbf{Instruction 1} and \textbf{Instruction 2} from Mind2Web are concerning about a \textit{ticket booking} task upon the same \textit{Event} domain and the same \textit{TicketCenter} website, which can be naturally combined into a natural conversation session. 

\paragraph{2) Decompose Complex Instructions}
Some instructions in Mind2Web exhibit complex action sequences, which are not common in daily conversations. 
On the other hand, complex instructions can serve as a good starting point for constructing follow-up instructions in multi-turn interactions. 
To facilitate the decomposition of complex instructions, we employ human-AI collaborative annotation, since AI is more proficient in determining how long action sequences can be divided into multiple executable sub-sequences while humans can decompose the instruction into multi-turn instructions in a more natural way. 
Specifically, we first employ ChatGPT for dividing the original instruction with complex action sequences into $N$ subtasks with corresponding action sub-sequences. Note that we set the target number of subtasks as $N=\lceil N'/4 \rceil$, where $N'$ is the number of actions in the original instruction. 
The prompt for instructing ChatGPT to decompose action sequences is as follows:
\begin{table}[h]
\setlength{\abovecaptionskip}{5pt}   
\setlength{\belowcaptionskip}{0pt}
    \centering
    \begin{tabular}{p{0.47\textwidth}}
    \hline
    \footnotesize\texttt{Analyze the instruction and corresponding actions provided for <domain> website, organize these actions into <N> distinct steps.} \\
    \hline
    \footnotesize\texttt{\#\#\# Requirements} \\
    \footnotesize\texttt{1. Review the instruction and related actions for completing a task on the specified website.} \\
    \footnotesize\texttt{2. Divide actions into logical, sequential steps.} \\
    \footnotesize\texttt{3. Format your response as a JSON array, with each object labeled as "step $i$" and containing an array of the sequential numbers of the actions that belong to each step.} \\
    \hline
    \footnotesize\texttt{\#\#\# Example} \\
    \footnotesize\texttt{\{ "step 1": [1, 2, 3], "step 2": [...], ... \}} \\
    \hline
    \footnotesize\texttt{\#\#\# Instruction}  \\\footnotesize\texttt{<original instruction>} \\
    \footnotesize\texttt{\#\#\# Actions} \\  \footnotesize\texttt{<original action sequences>} \\
    \hline
    \end{tabular}
    \label{tab:prompt_decompose}
\end{table}

As the example in Figure \ref{fig:pipeline}, the \textbf{Action Sequence 1} is sequentially decomposed into two action sub-sequences, including \textbf{Action Sub-sequence 1-1} and \textbf{Action Sub-sequence 1-2}. 
Then human annotators are asked to verify whether these sub-tasks are reasonable and executable. 
If not, they can re-arrange the decomposition based on their experiences from navigating the webpages. Overall, the pass rate of ChatGPT in decomposing action sequences is 98.5\%. 

\paragraph{3) Rewrite Conversational Instructions}
We refine the original standalone instructions into conversational ones by using anaphora and ellipsis, especially when consecutive instructions within a conversation session involve the same entities or the same actions. 
For example, \textbf{T2} uses \textit{one} to refer to the \textit{WWE ticket} mentioned in \textbf{T1}. 
While \textbf{T3} shifts to another task with the same action of \textit{booking tickets}, the verb $book$ is omitted. 
Similarly, the repeated content in \textbf{T3} is also omitted in \textbf{T4}. 

\paragraph{Quality Verification}
To ensure the quality of annotation in MT-Mind2Web, we conduct quality verification to validate the constructed conversations. 
If any mistake or problem is found, \textit{e.g.}, the constructed conversation is incoherent, the annotator will be asked to fix it until the annotation passes the verification. 

\subsection{Dataset Statistics}
After the dataset creation, we obtain a total of 720 web navigation conversation sessions, which contain 3,525 corresponding instruction and action sequence pairs in total and an average of 5 turns of user-agent interactions in each conversation session. Following the evaluation settings in Mind2Web \cite{mind2web}, we also select and divide the test set into three subsets, including cross-task, cross-website, and cross-subdomain, for evaluating how well an agent can generalize across tasks, websites, and domains. In specific, we select 44 samples for cross-subdomain evaluation from "Digital" and "Hotel", 42 samples for cross-website evaluation from "redbox", "viator", "nfl", "exploretock", "rentalcars", "cabelas", "bookdepository”, and 34 samples for cross-task evaluation. 
Then the remaining 600 samples are adopted as the training set. 
We present the train/test split in Table \ref{tab:data}.  Compared to traditional web navigation and conversational tasks, the conversational history can be extremely longer, including both the multi-turn user-agent conversation history and the multi-turn agent-environment interaction history within each conversation turn.

\begin{table}[t]
\setlength{\abovecaptionskip}{5pt}   
\setlength{\belowcaptionskip}{0pt}
    \centering
    \setlength{\tabcolsep}{1.5mm}{
    \begin{adjustbox}{max width=0.47\textwidth}
    \begin{tabular}{lrrrr}
    \toprule
    & \multirow{2}{*}{Train} & \multicolumn{3}{c}{Test (Cross-X)}  \\
    \cmidrule(lr){3-5}
    & & Task & Website & Subdomain \\
    \midrule
    \# Conversations  & 600 & 34 & 42 & 44 \\
    \# Turns & 2,896 & 191 & 218 & 216  \\
    Avg. \# Turn/Conv. & 4.83 & 5.62 & 5.19 & 4.91 \\
    Avg. \# Action/Turn & 2.95 & 3.16 & 3.01 & 3.07 \\
    Avg. \# Element/Turn & 573.8 & 626.3 & 620.6 & 759.4 \\
    Avg. Inst. Length & 36.3 & 37.4 & 39.8 & 36.2 \\
    Avg. HTML Length & 169K & 195K & 138K & 397K  \\
    \bottomrule
    \end{tabular}
    \end{adjustbox}}
    \caption{Statistics of the MT-Mind2Web dataset.}
    \label{tab:data}
    \vspace{-0.3cm}
\end{table}

\subsection{Problem Definition}
We introduce the task of \textbf{Conversational Web Navigation}, where the agent engages in not only multi-turn interactions with the environment, but also conversational interactions with the user. 
Given the conversational interaction history $C_t=\{q_1,A_1,...,A_{t-1},q_t\}$ where $A_i=\{a_i^1,a_i^2,...,a_i^k\}$ denotes the environment interaction history at each conversation turn, and the current environment state $E_t$ (\textit{e.g.}, HTML of the current webpage), the objective is to accurately predict the action sequence $A_t$ to accomplish the current user instruction $q_t$, which encompasses the target element for interaction and the operation. 

\begin{figure}[t]
\setlength{\abovecaptionskip}{5pt}   
\setlength{\belowcaptionskip}{0pt}
\centering 
\includegraphics[width=0.48\textwidth]{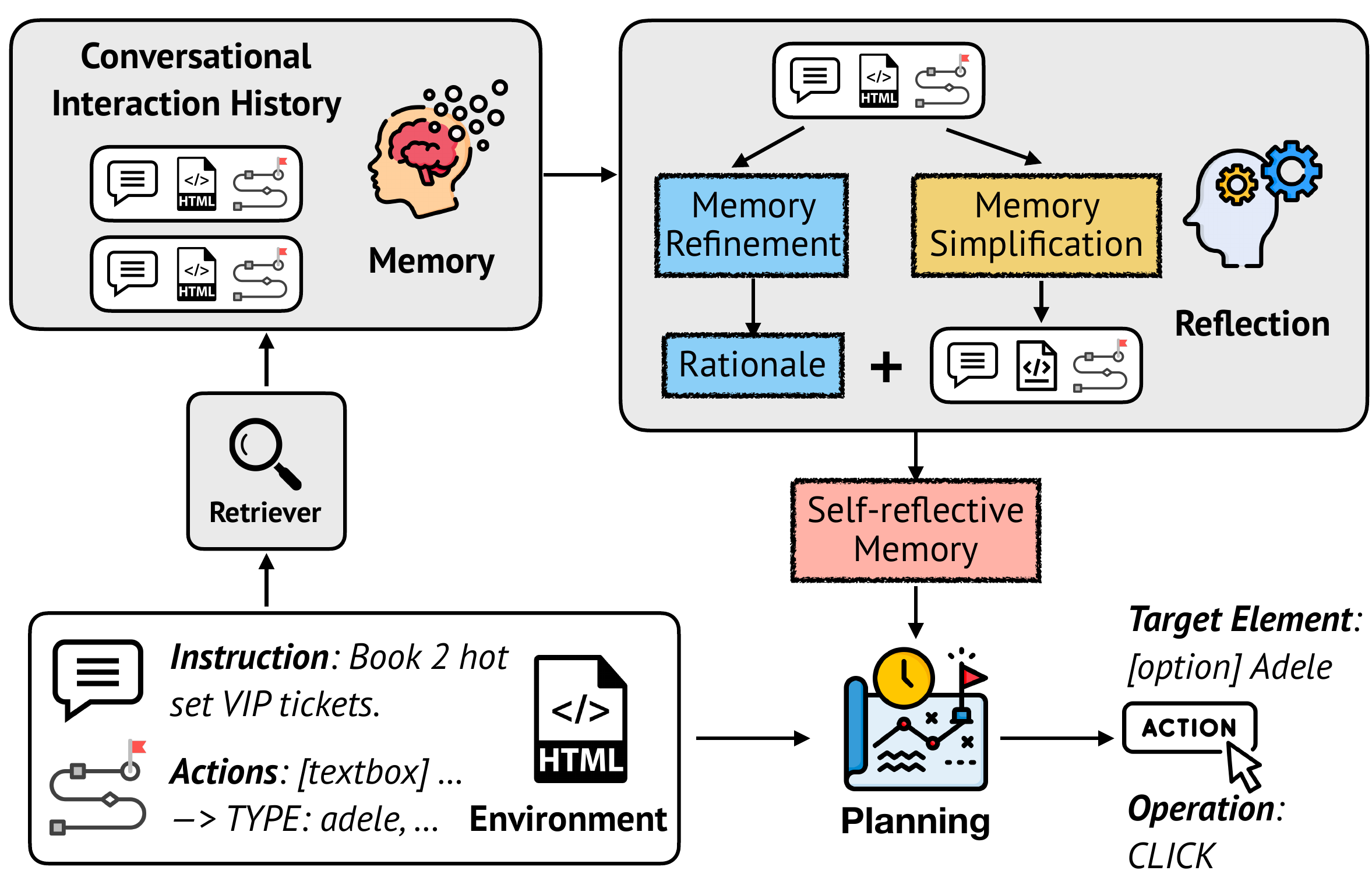}
\caption{Overview of Self-MAP.}
\label{fig:method}
\vspace{-0.3cm}
\end{figure}

\section{Method}
We introduce the Self-MAP framework, which combines memory-augmented planning with self-reflection. The overview of Self-MAP is presented in Figure \ref{fig:method}, consisting of three main components: Memory, Reflection, and Planning Modules.

\subsection{Memory Module}
The memory bank for conversational web agents is constructed by the conversational interaction history $C_t$, where each memory snippet can be represented by $M_t^k=\{q_t,A_t^{k-1},E_t^k,a_t^k\}$.  
It requires a significant number of tokens to inject each memory snippet into the current running memory of the agent, which will be limited by the maximum input length of language models. 
Meanwhile, some memory snippets, due to the irrelevance and inconsistency of their instructions and actions to the current environment setting, fail to provide useful guidance for the agent to predict the subsequent action. 
As such, we introduce a multifaceted matching approach to retrieve the top-$K$ relevant snippets within the memory bank at the action level. 

%User conversation history naturally includes gold-standard examples and ample dialog contexts, serving as a reference point for agents. However, the task of describing each interaction turn within environments and the behaviors of agents requires a significant amount of tokens, limited by the maximum context length of language models. We observed that some history instructions and actions, due to its irrelevance and inconsistency with the current environment setting, fail to provide guidance for the agent to predict the subsequent action. 

Formally, given an on-going conversational interaction trajectory $C_t^k = \{q_1, A_1, \ldots, q_t, A_t^{k-1}\}$, where $A_t^{k-1} = \{a_t^1, a_t^2, \ldots, a_t^{k-1}\}$ represents the trajectory of agent-environment interactions at the current conversation turn, multifaceted matching  constructs the query using both the user instruction and the present agent action sequence $(q_t, A_t^{k-1})$ to retrieve relevant memory snippets from the memory bank. In this manner, the query encodes not only the semantic relevance of the current instruction to the conversation context but also the similarity of the action trajectory to the historical interactions. 
Specifically, we adopt OpenAI's \texttt{text-embedding-ada-002} as the embedding method to transform the query and the memory snippets into vector representations. Then we compute the cosine similarity in the embedding space for retrieving top-$K$ memory snippets.

\subsection{Reflection Module}
Due to the limitation of on-going memory space (\textit{i.e.}, input length limitation) for LLM-powered agents, we design a reflection module to maximize the utility of the limited memory space, which involves two steps: 1) Memory Simplification, and 2) Memory Refinement. 

\paragraph{Memory Simplification}
In the candidate generation process in the \textsc{MindAct} framework \cite{mind2web}, a small pre-trained LM (\textit{e.g.}, DeBERTa \cite{deberta}) is adopted for ranking the top-$N$ candidate DOM elements from the environment state (\textit{i.e.}, HTML) that are related to the instruction and the current step for improving the final action prediction. 
To simplify each memory snippet, we apply the same process to remove task-irrelevant and noisy elements from the environment state, thereby freeing up memory space for more extensive conversation history retention. 
Afterwards, we denote the simplified environmental state $E^k_t$ in the memory snippet $M^k_t$ as $e^k_t$.

\paragraph{Memory Refinement}

Drawing inspiration from self-reflection techniques \cite{reflexion,self-rag}, we design a specialized Memory Refinement approach for the domain of conversational web navigation. This module diverges from traditional self-reflection methods, as it does not collect incorrect trajectories for the model to analyze. This is primarily due to the constraints of a static evaluation setting and the limited context length to present the full webpage. Instead, we leverage the exceptional reasoning capability of LLMs to generate intermediate reasoning rationale as a supervised signal to enrich the memory information. For each retrieved memory snippet $(q_t, A_t^{k-1}, a_t^k)$, we prompt the LLM to generate an in-depth rationale $r_t^k$ explaining the reason for the decision-making process of the next action.

\begin{table*}[t]
\setlength{\abovecaptionskip}{5pt}   
\setlength{\belowcaptionskip}{0pt}
\centering
\label{tab:overall}
\setlength{\tabcolsep}{1mm}{
    \begin{adjustbox}{max width=\textwidth}
\begin{tabular}{lcccccccccccc}
\toprule
 & \multicolumn{4}{c}{Cross-Task} & \multicolumn{4}{c}{Cross-Website} & \multicolumn{4}{c}{Cross-Subdomain} \\
\cmidrule(lr){2-5}\cmidrule(lr){6-9}\cmidrule(lr){10-13}
 & Ele. Acc & Op. F1 & SSR & TSR & Ele. Acc & Op. F1 & SSR & TSR & Ele. Acc & Op. F1 & SSR & TSR \\
\midrule
DeBERTa \cite{deberta} & 36.8 & - & - & - & 31.7 & - & - & - & 27.7 & - & - & - \\
\textsc{MindAct} (GPT-3.5) \cite{mind2web} & 4.3 & 27.6 & 1.9 & 1.0 & 6.7 & 22.2 & 2.1 & 1.7 & 4.0 & 22.9 & 1.5 & 1.1 \\
\midrule
    \textsc{MindAct} (Flan-T5$_{\text{base}}$)  \cite{mind2web}  & 43.2 & 79.1 & 36.6 & 14.2 & 38.8 & 69.4 & 29.2 & 15.2 & 41.9 & 77.2 & 35.5 & 15.7 \\
\textsc{MindAct} + CAR  \cite{qr} & 47.8 & 78.8 & 41.4 & 16.1 & 37.0 & 67.5 & 32.2 & 9.6 & 41.2 & 75.3 & 35.4 & 13.2 \\
\textsc{MindAct} + Fixed \cite{fixed} & 51.0 & 80.8 & 42.6 & 18.4 & 42.4 & 70.0 & 35.4 & 15.3 & 43.1 & 77.6 & 37.5 & 17.7 \\
Synapse \cite{iclr24-trajectory} & 49.6 & 79.9 & 41.9 & 18.4 & 43.1 & 70.6 & 33.1 & 13.7 & 41.7 & 77.8 & 35.9 & 16.0 \\ 
\textbf{Self-MAP} & \textbf{56.2} & \textbf{82.5} & \textbf{47.1} & \textbf{24.7} & \textbf{48.3} & \textbf{71.8} & \textbf{40.6} & \textbf{18.2} & \textbf{46.4} & \textbf{79.1} & \textbf{38.3} & \textbf{20.8}\\
\midrule
\textsc{MindAct} (Flan-T5$_{\text{large}}$)  \cite{mind2web} &  \textbf{59.0} & 80.6 & \textbf{53.2} & 26.0 & 43.6 & 67.6 & 36.5 & 12.4 & 46.8 & 74.0 & 38.9 & 21.8\\
\textsc{MindAct} + CAR \cite{qr} & 54.5 & 79.5 & 47.8 & 19.8 & 43.2 & 69.2 & 36.1 & 12.2 & 44.5 & 75.0 & 40.2 & 15.6\\
\textsc{MindAct} + Fixed \cite{fixed} & 58.0 & 79.7 & 51.3 & 26.4 & \textbf{46.2} & \textbf{69.7} & \textbf{37.6} & 15.2 & 47.4 & 74.9 & 38.8 & 21.4   \\
Synapse \cite{iclr24-trajectory} & 57.5 & \textbf{82.0} & 50.0 & 23.2 & 45.1 & 69.0 & 37.1 & 13.0 & 47.4 & 74.1 & 39.3 & 19.4 \\
\textbf{Self-MAP} & 58.1 & 80.5 & 51.7 & \textbf{26.6} & 44.8 & 68.8 & 36.8 & \textbf{15.7} & \textbf{52.0} & \textbf{77.1}  & \textbf{43.6} & \textbf{25.4} \\
\bottomrule
\end{tabular}
\end{adjustbox}}
\caption{Experimental results on MT-Mind2Web. TSR can be regarded as the main metric.}
\label{table:overall}
\vspace{-0.3cm}
\end{table*}

\paragraph{Self-reflective Memory}
After the previous two steps, we obtain the self-reflective memory snippet, which not only filters out the irrelevant and noisy information from the environmental state but also integrates the additional informative rationale. We denote the self-reflective memory snippet as $\hat{M}_t^k=\{q_t,A_t^{k-1},e_t^k,a_t^k,r_t^k\}$. 

\subsection{Planning with Self-reflective Memory}
For each interaction step $k$ at the current conversation turn $t$, given the current user instruction $q_t$ and previous action sequences $A_t^{k-1}$, we first obtain the top-$K$ retrieved memory snippets with self-reflection $\mathcal{M}_t^k=\{\hat{M}\}^K$ from the reflection module, and the top-$N$ candidate elements $e^k_t$ simplified from the current environment state $E^k_t$ using the same ranker as memory simplification. 
Then we fine-tune the LLM to plan the next action $a_t^k$ including the target element and the operation, based on the input consisting of $(q_t,A_t^{k-1},e^k_t,\mathcal{M}_t^k)$. 
Similar to the settings in \citet{mind2web}, there are two types of planning paradigms: 1) Multi-choice Question Answering, and 2) Direct Generation.

\begin{table*}[t]
\setlength{\abovecaptionskip}{5pt}   
\setlength{\belowcaptionskip}{0pt}
\centering
\setlength{\tabcolsep}{1mm}{
    \begin{adjustbox}{max width=\textwidth}
\begin{tabular}{lcccccccccccc}
\toprule
 & \multicolumn{4}{c}{Cross-Task} & \multicolumn{4}{c}{Cross-Website} & \multicolumn{4}{c}{Cross-Subdomain} \\
\cmidrule(lr){2-5}\cmidrule(lr){6-9}\cmidrule(lr){10-13}
 & Ele. Acc & Op. F1 & SSR & TSR & Ele. Acc & Op. F1 & SSR & TSR & Ele. Acc & Op. F1 & SSR & TSR \\
\midrule
\textbf{Self-MAP} & \textbf{56.2} & \textbf{82.5} & \textbf{47.1} & \textbf{24.7} & 48.3 & \textbf{71.8} & \textbf{40.6} & \textbf{18.2} & \textbf{46.4} & 79.1 & \textbf{38.3} & \textbf{20.8}\\

w/o Generation-based Planning & 51.7 & 79.4 & 43.5 & 22.2 & 43.1 & 69.5 & 34.9 & 15.5 & 44.8 & 77.2 & 37.3 & 17.7 \\
w/o Memory Simplification & 50.5 & 80.7 & 41.0 & 20.7 & 44.9 & 69.6 & 36.9 & 16.6 & 42.3 & \textbf{79.2} & 36.4 & 15.9 \\
w/o Memory Refinement & 52.1 & 81.3 & 43.0 & 23.2 & \textbf{48.9} & 70.8 & 39.1 & 18.1 & 46.3 & 78.7 & 37.2 & 17.8\\
w/o Multifaceted Matching & 52.6 & 80.6 & 44.3 & 21.6 & 46.9 & 71.2 & 37.9 & 17.2 & 44.8 & 78.6 & 35.8 & 17.8 \\
\bottomrule
\end{tabular}
\end{adjustbox}}
\caption{Ablation study. "w/o Generation-based Planning" denotes that we use MCQ-based Planning, while "w/o Multifaceted Matching" denotes that we prepend the chronological conversation context without retrieval.}
\label{tab:ablation}
\vspace{-0.3cm}
\end{table*}

\section{Experiment}
\subsection{Experimental Setups}
\paragraph{Baselines}
As conversational web navigation is a new task, we first adapt several state-of-the-art traditional web navigation methods as baselines, including DeBERTa \cite{deberta}, \textsc{MindAct} \cite{mind2web}, \textsc{MindAct} + Fixed \cite{fixed}, and Synapse \cite{iclr24-trajectory}. We further include a classic baseline for conversational tasks, \textit{i.e.}, \textsc{MindAct} + Context-Aware Rewriting (CAR) \cite{qr}. Details of these baselines are presented in Appendix \ref{app:baseline}.

\paragraph{Evaluation Metrics}
Following the single-turn setting \cite{mind2web}, we adopt the following metrics for evaluation: 1) Element Accuracy (Ele. Acc) matches the selected element with all required elements. 2) Operation F1 (Op. F1) stands for the token-level F1 score for the predicted operation. 3) Step Success Rate (SSR). An interaction step is regarded as successful only if both the selected element and the predicted operation are correct. 4) Turn Success Rate (TSR). A conversation turn is regarded as successful only if all steps at this turn have succeeded. We report macro average metrics, which first calculate the average per task, and then average over all tasks.

\paragraph{Implementation Details}
The overall Self-MAP framework basically follows the same configuration as \textsc{MindAct} for a fair comparison. 
Specifically, we use the base version of DeBERTa-v3 \cite{deberta} as the candidate HTML element ranker. We adopt the base and large versions of Flan-T5 \cite{flant5} as the generation model to plan the next action. All the usage of ChatGPT in the experiments is based on \texttt{gpt-3.5-turbo-1106}. Flan-T5$_\text{base}$ and Flan-T5$_\text{large}$ are trained on servers with 4 A5000 24GB GPUs. DeBERTa is trained with single A100 40GB GPU. More implementation details are presented in Appendix \ref{app:implementation}.

\subsection{Overall Evaluation}\label{subsec:overall_evaluation}
Experimental results on MT-Mind2Web are summarized in Table \ref{table:overall}. 
Among the baselines, similar to the findings in \citet{mind2web}, DeBERTa, which only performs element selection, and \textsc{MindAct} (GPT-3.5), which relies on the in-context learning capabilities of LLMs without fine-tuning, fall short of tackling the web navigation problem. 
For \textsc{MindAct}+CAR, we observe that its performance is worse than the vanilla \textsc{MindAct} (except for Cross-Task with Flan-T5$_\text{base}$), where GPT-3.5 fails to effectively rewrite the current conversational instruction, which further obfuscates the original instruction. 
In contrast, both \textsc{MindAct}+Fixed and Synapse generally outperform  \textsc{MindAct}, which also validates our motivation of retrieving memory from the conversational interaction history. 
Surprisingly, Synapse (SOTA method in Mind2Web) performs even worse than \textsc{MindAct}+Fixed which employs the fixed memory selection. This  indicates the coarse-grained $k$NN matching in Synapse fails to effectively measure the relevance between the current conversation status and the candidate memory snippets in our conversational setting. 
In general, using a stronger base model (\textit{e.g.}, Flan-T5$_\text{large}$) improves the final performance. 
Overall, Self-MAP consistently and substantially outperforms these baselines with a noticeable margin (\textit{e.g.}, +6.3/+2.9/+3.1 TSR scores with Flan-T5$_\text{base}$ over the strongest baselines). 
This showcases the effectiveness of utilizing our proposed memory-augmented planning framework as well as the self-reflection strategy for enhancing memory utilization.

\subsection{Ablation Study}
To validate the specific designs of the Self-MAP framework, we present the ablation study in Table \ref{tab:ablation}. First, we observe that Generation-based Planning substantially surpasses MCQ-based Planning in performance. This superiority is attributed not only to the advanced generative capabilities of large language models (LLMs) but also to their efficiency in conserving context space for memory utilization.
Second, the process of Memory Simplification emerges as the most critical factor in enhancing overall performance. This finding underscores the importance of optimizing the use of limited context space, highlighting the necessity of efficient memory management.
Third, the contribution of Memory Refinement is notably more pronounced in cross-task scenarios compared to other settings. This indicates its relatively lower generalizability in modeling decision-making processes, compared to the other components of our framework.
Lastly, Multifaceted Matching for memory retrieval significantly outperforms vanilla memory prepending, which suggests the necessity of filtering out noisy conversational interaction history to focus on the relevant part.

\begin{figure}[t]
\setlength{\abovecaptionskip}{5pt}   
\setlength{\belowcaptionskip}{0pt}
\centering 
\includegraphics[width=0.48\textwidth]{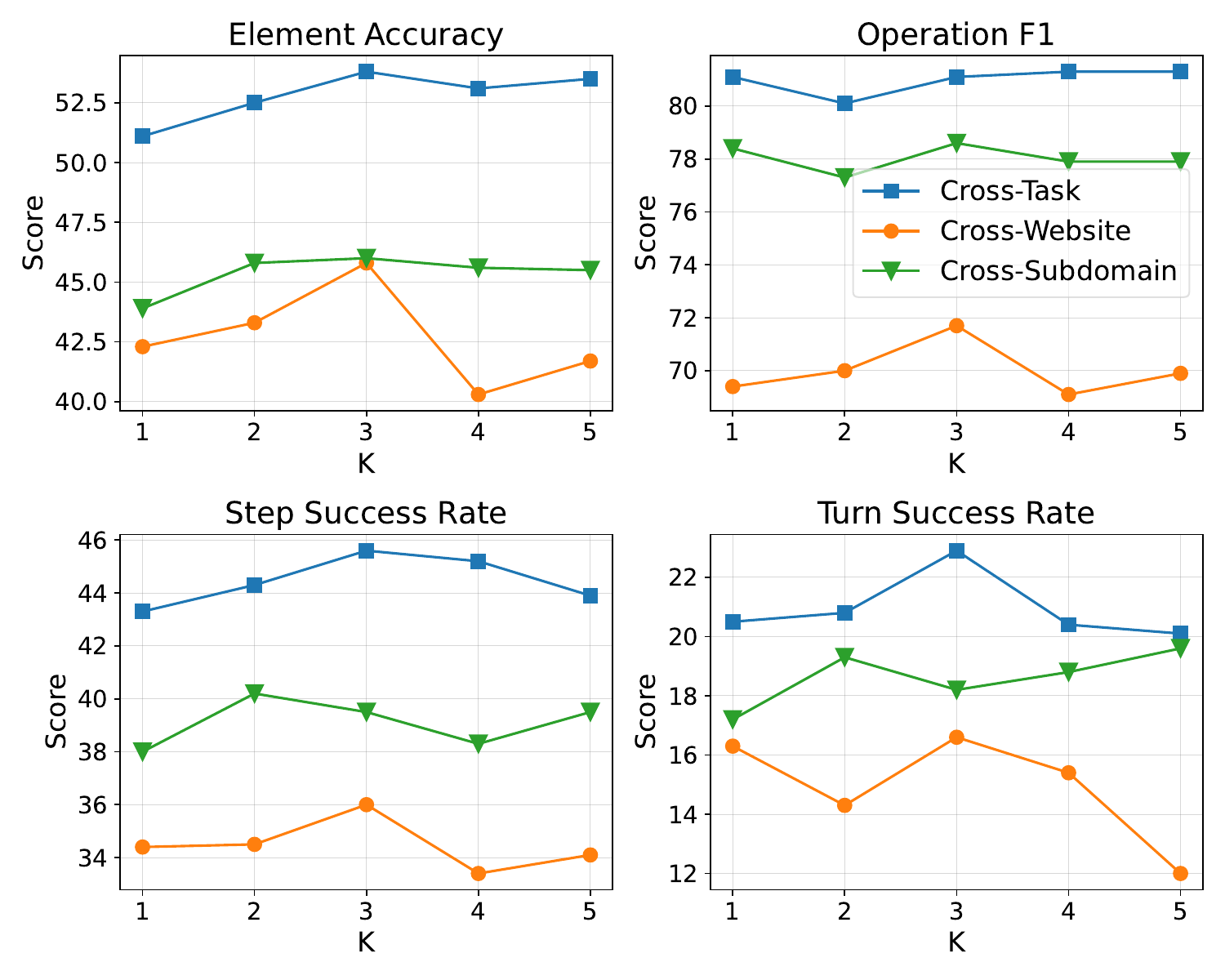}
\caption{Performance in terms of different number of retrieved memory snippets.}
\label{fig:range_k}
\vspace{-0.3cm}
\end{figure}

\begin{figure*}
\setlength{\abovecaptionskip}{0pt}   
\setlength{\belowcaptionskip}{0pt}
    \centering
    \includegraphics[width=\textwidth]{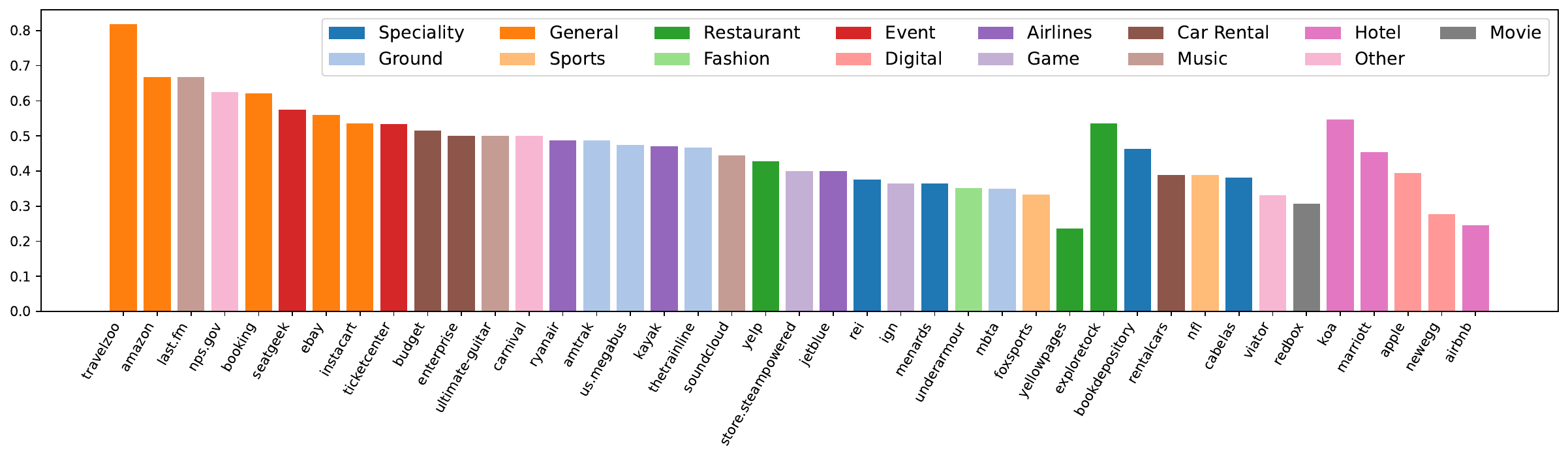}
    \caption{Step success rate regarding each website grouped by the three test splits. 
    }
    \label{fig:generalization}
    \vspace{-0.3cm}
\end{figure*}

\begin{figure}[t]
\setlength{\abovecaptionskip}{5pt}   
\setlength{\belowcaptionskip}{0pt}
\centering 
\includegraphics[width=0.48\textwidth]{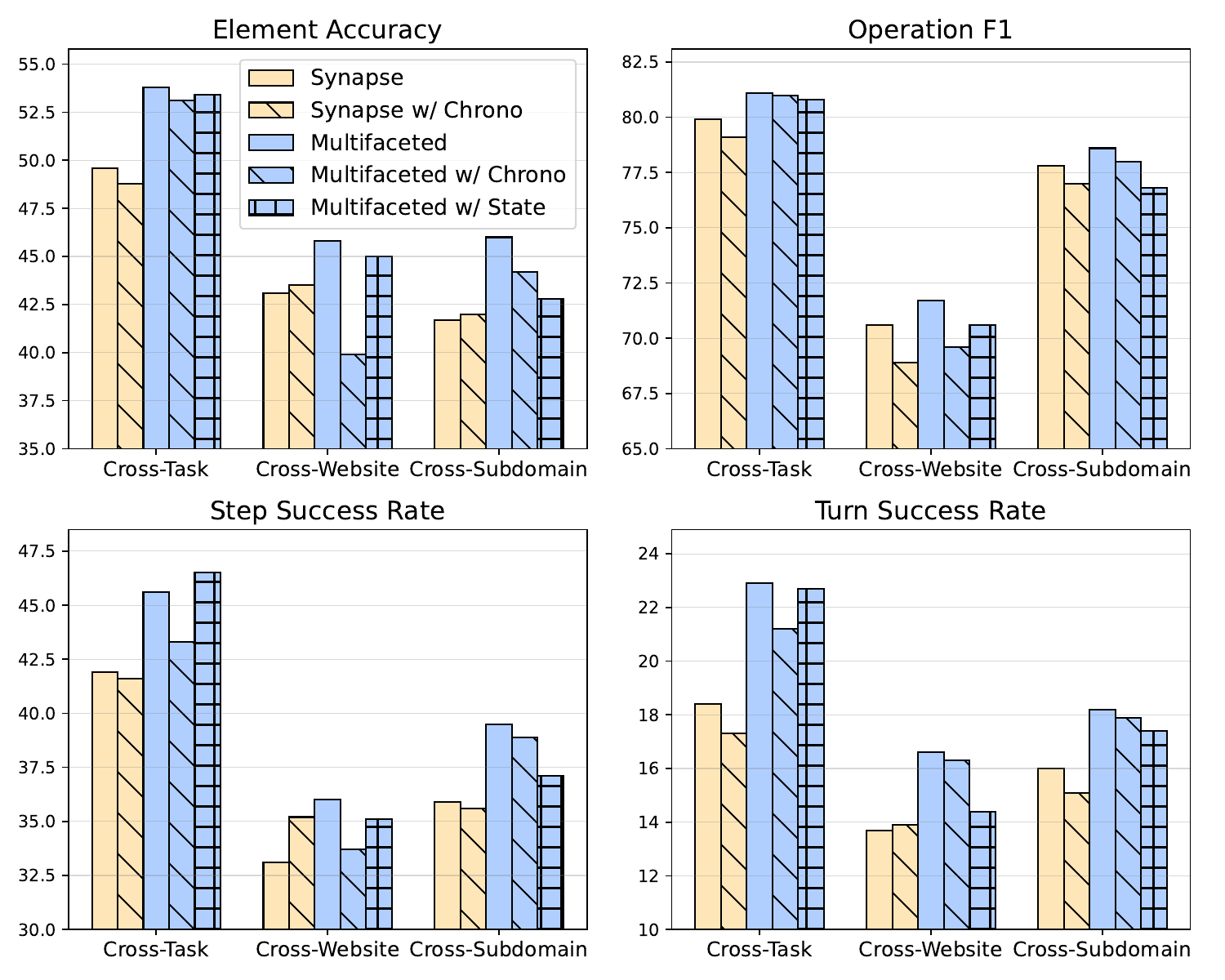}
\caption{Performance in terms of different number of retrieved memory snippets.}
\label{fig:prompt_study}
\vspace{-0.3cm}
\end{figure}

\subsection{Detailed Analysis}

\paragraph{Effect of the Number of Retrieved Memory}
We first analyze the effect of the number of retrieved memory snippets by varying $K$ from 1 to 5. The results are presented in Figure \ref{fig:range_k}. 
We observe that the performance increases along with the growth of the number of retrieved memory snippets at the beginning ($K\leq 3$), indicating the value of refining the memory utility for exploiting more relevant information. 
However, the continued increase on $K$ fails to contribute to the performance improvement, even making worse performance in some subsets, (\textit{e.g.}, cross-task and cross-website). 
As shown in the dataset statistics (Table \ref{tab:data}), the average number of conversational turns is about 5 turns. Therefore, it may introduce noisy information from those irrelevant turns when increasing the number of retrieved memory snippets.

\paragraph{Analysis of Generalizability}
Compared with the analysis of generalizability conducted in Mind2Web \cite{mind2web}, we draw some similar observations from Figure \ref{fig:generalization} in the conversational setting: 1) All models perform better on the Cross-Task setting than the other two settings. 2) There is no significant difference between Cross-Website and Cross-Subdomain settings, indicating that the challenges primarily come from the diversity in website designs and interaction logic rather than domain specifics. 
Differently, we observe that the performance gap between the Cross-Task setting and the other two settings is more substantial than that in Mind2Web (10\%$\rightarrow$20\%), which suggests that the interaction logic becomes more complicated when introducing multi-turn user-agent interactions. 

\paragraph{Analysis of Conversation Prompt Designs} 

Unlike traditional web navigation tasks, more types of information are supposed to be included in the prompt for action planning in the MT-Mind2Web task.
We first examine the impact of memory snippet order in conversation prompts, including the adopted relevance-based order and the typical chronological (sequential) order, in the Synapse and Self-MAP methods.
As shown in Figure \ref{fig:prompt_study}, both methods generally perform much better with relevance-based order compared to chronological order. 
In addition, we introduce state-based information into the proposed multifaceted matching approach. In Self-MAP, we omit $A_t^{k-1}$ in the $M_t^k$, as in actual conversational contexts, explicitly identifying the state within a sequence-ordered trajectory is unnecessary. However, in the context of action-level matching, which lacks a sequential framework, state-based information cannot be inferred from the trajectory. Our results suggest that multifaceted matching typically achieves better performance without state-based information in the retrieved memory. 
Based on these analyses, we finalize our prompt designs, which are presented in Appendix \ref{app:planning}.

\section{Conclusions}
To investigate the capability of web agents to follow instructions over multiple turns, we introduce the MT-Mind2Web dataset for conversational web navigation, which requires complex, multi-turn interactions with both users and the web environment. To overcome the underlying challenges, such as the restricted context length of LLMs and their dependency on conversational context, we present a novel framework named Self-Reflective Memory-Augmented Planning (Self-MAP), which utilizes memory augmentation and self-reflection techniques. We rigorously evaluate the MT-Mind2Web dataset against extensive baselines, conducting thorough analyses across various domains. Our experimental findings demonstrate the effectiveness of our proposed approach.

%\clearpage
\section*{Limitation}

\paragraph{Multimodal Environment}
With the advent of multimodal LLMs, recent studies demonstrate the effectiveness of applying multimodal web agents \cite{seeact,webvoyager} onto the web navigation problem. Without loss of generality, the constructed MT-Mind2Web dataset can also be adapted to the multimodal environment as the original Mind2Web dataset. In this work, we mainly focus on benchmarking general HTML-grounded methods, while we believe that it will also be a promising research direction on studying the conversational web navigation problem under the multimodal setting. 

\paragraph{Online Evaluation}
As a pioneer study of conversational web agents, we follow the typical offline evaluation settings of both conversational tasks \cite{nips23-mtbench} and single-turn web navigation tasks \cite{mind2web}, which allows researchers and practitioners to efficiently and conveniently evaluate the web gents using snapshots of complex real-world websites. However, it also inherits the drawback of the offline evaluation setting, \textit{e.g.}, evaluating dynamic interactions. 

\bibliography{custom}

\appendix
\section*{Appendix}
\section{Details of Experimental Setups}
\subsection{Details of Baselines}
\label{app:baseline}
\begin{itemize}[leftmargin=*]
    \item DeBERTa \cite{deberta}. Following \citet{mind2web}, we also fine-tune DeBERTa as the ranker for selecting target elements. 
    \item \textsc{MindAct} \cite{mind2web} performs multi-choice question answering to select the target element from a list of options. Under the conversational setting, the input includes the whole conversational interaction history. 
    \item \textsc{MindAct} + CAR \cite{qr}. We first employ context-aware rewriting (CAR) using ChatGPT to reconstruct the self-contained instructions from the conversational instructions and the conversation context. Then the self-contained instructions are directly used as the input instructions for Mind2Act. The prompting details are presented in Appendix \ref{app:rewrite}. 
    \item \textsc{MindAct} + Fixed \cite{fixed}. \citet{fixed} empirically observe that using fixed examples outperforms relevance-based example selection for demonstration-based learning in the web navigation task. We fix the first 3 turns in the conversation history in chronological order as the memory. 
    \item Synapse \cite{iclr24-trajectory}. Synapse employs metadata, including website, domain, subdomain, and task as keywords to conduct $k$NN-based exemplar retrieval. Given that each conversation turn in our task shares the same website, domain, and subdomain information, we only keep the task in the metadata and perform the turn-level $k$NN. 
\end{itemize}

\subsection{More Details on Implementation}\label{app:implementation}

\input{prompts/rewrite}

\begin{itemize}[leftmargin=*]
    \item \textbf{Memory Simplification}. We use Sentence-Transformers\footnote{\url{https://www.sbert.net/examples/applications/cross-encoder/README.html}} and fine-tune DeBERTa-v3-base \cite{deberta} for our multi-turn task. Following \citet{mind2web}, we choose 5 random elements, including the positive candidate for training, and select the top-50 elements compared in groups of 5 for evaluation. During the training, we set the batch size as 32, the learning rate as 3e-5, and trained for 5 epochs. 
    \item \textbf{Action Planning}. We use Flan-T5$_\text{base}$ and Flan-T5$_\text{large}$ \cite{flant5} for MCQ-based and generation-based action planning. We set the maximum sequence length at 2,048. 
    Since the max context length for the tokenizer is 512, we tokenize the system message, HTML, user input, and assistant response separately. 
    During the training, we set the batch size as 8 and 4 for Flan-T5$_\text{base}$ and Flan-T5$_\text{large}$ respectively, the learning rate as 5e-5, and trained for 5 epochs. 
    \item \textbf{Multifaceted Matching}. We use the OpenAI embedding model \texttt{text-embedding-ada-002} for matching, and choose cosine similarity for calculating embedding.
    We set the number of retrieved memories $K$ to 3.  
    The prompting details of two paradigms of action planning are introduced in Appendix \ref{app:planning}.
    \item \textbf{Memory Refinement} We use ChatGPT with the version of \texttt{gpt-3.5-turbo-1106}, maximum new tokens as 100, and temperature as 0. We only extract HTML snippets of the positive element for ChatGPT to generate rationales. If there is no positive element in the HTML snippet, we use "The assistant's answer is derived from the absence of a specific option in the provided HTML content, leading to the conclusion that none of the options provided are suitable for the user's task." as the default rationale. The prompting details for memory refinement are presented in Appendix \ref{app:reflexion}. 
\end{itemize}

\section{Prompting Details}
\subsection{Prompts for Context-aware Rewriting}\label{app:rewrite}
The prompts for implementing Context-Aware Rewriting (CAR) \cite{qr} are presented in Table \ref{table:rewrite_prompt}. 

\subsection{Prompts for Planning}\label{app:planning}
The prompt templates for Flan-T5 planning are presented in Table \ref{table:t5_mcq_template} for MCQ-based planning and in Table \ref{table:t5_generation_template} for generation-based planning. Note that \texttt{\{\}} represents predefined placeholders in the MT-Mind2Web task. 
The system message for GPT-3.5 planning is listed in Table \ref{table:gpt_system_message}.

\input{prompts/t5}
\input{prompts/gpt}

\subsection{Prompts for Memory Refinement}\label{app:reflexion}
The prompts for implementing memory refinement using GPT-3.5 are shown in Table \ref{table:reflection_prompt}.
\input{prompts/reflexion}

\section{Case Study}
\subsection{Conversation Description}
This section examines a specific case from the MT-Mind2Web task, focusing on the responses from \textsc{MindAct} + CAR, Synapse, and Self-MAP.
The conversation history, comprising seven trajectories, is detailed in Table \ref{table:case_conversation_history}.
The instruction of the current conversation is \textit{Search \textquotesingle xbox series x console\textquotesingle.}, with no previous actions noted.
\subsection{Analysis and Comparison}
The analysis of different final evaluation steps for the current conversation is shown in Table \ref{table:case_last_conversation}. For \textsc{MindAct} + CAR, we highlight the rephrased query in blue. This model, however, integrates irrelevant turns from the conversation history, aligning with our observations in Subsection \ref{subsec:overall_evaluation}.

Synapse employs a coarse-grained $k$NN matching method, retaining all historical conversation turns.
Compared with Synapse, Table \ref{table:case_history_trajectory} displays the augmented memory and self-reflection from Self-MAP. Notably, Self-MAP selects Trajectories 2, 1, and 7 due to their relevance to the current instruction. These selections are along with reasoning rationales generated by GPT-3.5 and highlighted in blue. Both Synapse and Self-MAP select identical HTML elements in their final evaluation step, as indicated in Table \ref{table:case_last_conversation}. Synapse does not process the sequence of the search operation correctly. This oversight makes it trigger a submit action without entering the search keyword, as a result of the noisy information in its retrieved memory. Conversely, Self-MAP's success in the same scenario can be attributed to its understanding and combination of relevant conversation history from multifaceted matching and self-reflection, highlighting the efficacy of its approach.
\input{prompts/case_study}
\end{document}

%% file: prompts/rewrite.tex
\begin{table*}[t]
\small
\centering
\begin{tabularx}{\textwidth}{p{1.25cm}X}

    \toprule
    \multicolumn{1}{c}{\textbf{Role}} & \multicolumn{1}{c}{\textbf{Content}} \\
    \midrule
    system & You are a helpful assistant adept at understanding and rewriting user queries. Your task is to evaluate the relevance of previous queries, add any relevant missing details from the previous queries, and rewrite the current query. \\ \midrule
    user & Rewrite: Help me check the popularity in 2015.\newline
    Previous queries:\newline
    Find the baby girl's name.\newline
    Show me the most popular one. \\
    & \\
    assistant & Show me the popularity in 2015 of the current most popular baby girl name. \\ \midrule
    user & Rewrite: List the best rated for me.\newline
    Previous queries:\newline
    Search for pizza recipes for me.\newline
    I want the ones that take 30 minutes or less.\newline
    Show me the vegan option.\newline
    Find Halloween dishes.\newline
    Help me sort by rating.\newline
    Find pie recipes.\newline
    Show me all the content. \\
    & \\
    assistant & Find pie recipes and show the best rated ones. \\ \midrule
    user & Rewrite: How about a list of CDB product reviews.\newline
    Previous queries:\newline
    Find me a gluten-free diet to lose weight for a pregnant woman. \\
    & \\
    assistant & Browse a list of CDB product reviews. \\
    \bottomrule
\end{tabularx}
\caption{Prompts for context-aware rewriting. }
\label{table:rewrite_prompt}
\end{table*}

%% file: prompts/t5.tex
\begin{table*}[t]
\small
\centering
\begin{tabularx}{\textwidth}{X}
\toprule
\textbf{System Message} \\
You are a helpful assistant that is great at website design, navigation, and executing tasks for the user.\\[1ex] 
\midrule
\textbf{Conversation History} \\
\#\#\# Human: \textasciigrave\textasciigrave\textasciigrave\\
\texttt{\{HTML snippets including 5 elements\}} \\
\textasciigrave\textasciigrave\textasciigrave\\[1ex] 
Based on the HTML webpage above, try to complete the following task:\\
Task: \texttt{\{instruction\}}\\
What should be the next action? Please select from the following choices (If the correct action is not in the page above, please select A. 'None of the above'):\\\\
A. None of the above\\
B. \texttt{\{element 1\}}\\
C. \texttt{\{element 2\}}\\
D. \texttt{\{element 3\}}\\
E. \texttt{\{element 4\}}\\
F. \texttt{\{element 5\}}\\[1ex] 
\#\#\# Assistant: \texttt{\{response\}} \\
\texttt{\{Optional: Reflection\}}\\
\ldots \\
\midrule
\textbf{Current Conversation} \\
\#\#\# Human: \textasciigrave\textasciigrave\textasciigrave\\
\texttt{\{HTML snippets including 5 elements\}} \\
\textasciigrave\textasciigrave\textasciigrave\\[1ex] 
Based on the HTML webpage above, try to complete the following task:\\
Task: \texttt{\{instruction\}}\\
Previous actions:\\ \texttt{\{last 5 action representations\}} \\
What should be the next action? Please select from the following choices (If the correct action is not in the page above, please select A. 'None of the above'):\\\\
A. None of the above\\
B. \texttt{\{element 1\}}\\
C. \texttt{\{element 2\}}\\
D. \texttt{\{element 3\}}\\
E. \texttt{\{element 4\}}\\
F. \texttt{\{element 5\}}\\[1ex] 
\#\#\# Assistant: \texttt{\{response\}} \\
\bottomrule
\end{tabularx}
\caption{Prompt Templates for MCQ-based Flan-T5 Planning}
\label{table:t5_mcq_template}
\end{table*}

\begin{table*}[t]
\small
\centering
\begin{tabularx}{\textwidth}{X}
\toprule
\textbf{System Message} \\
You are a helpful assistant that is great at website design, navigation, and executing tasks for the user.\\[1ex] 
\midrule
\textbf{Conversation History} \\
\#\#\# Human: \textasciigrave\textasciigrave\textasciigrave\\
\texttt{\{HTML snippets including 5 elements\}} \\
\textasciigrave\textasciigrave\textasciigrave\\[1ex] 
Based on the HTML webpage above, try to complete the following task:\\
Task: \texttt{\{instruction\}}\\
What should be the next action? Please select the element to interact with, and the action to perform along with the value to type in or select. If the task cannot be completed, output None: \\[1ex] 
\#\#\# Assistant: \texttt{\{response\}} \\
\texttt{\{Optional: Reflection\}}\\
\ldots \\
\midrule
\textbf{Current Conversation} \\
\#\#\# Human: \textasciigrave\textasciigrave\textasciigrave\\
\texttt{\{HTML snippets including 5 elements\}} \\
\textasciigrave\textasciigrave\textasciigrave\\[1ex] 
Based on the HTML webpage above, try to complete the following task:\\
Task: \texttt{\{instruction\}}\\
Previous actions:\\ \texttt{\{last 5 action representations\}} \\
What should be the next action? Please select the element to interact with, and the action to perform along with the value to type in or select. If the task cannot be completed, output None: \\[1ex] 
\#\#\# Assistant: \texttt{\{response\}} \\
\bottomrule
\end{tabularx}
\caption{Prompt Templates for Generation-based Flan-T5 Planning}
\label{table:t5_generation_template}
\end{table*}

%% file: prompts/gpt.tex
\begin{table*}[t]
\small
\centering
\begin{tabularx}{\textwidth}{X}
\toprule
You are a helpful assistant that is great at website design, navigation, and executing tasks for the user.\\[1ex] 
\#\#\# Example\\ 
User:\\ 
\textasciigrave\textasciigrave\textasciigrave\\ 
<html> <div> <div> <a tock home page /> <button id=0 book a reservation. toggle open> <span> Book a reservation </span> </button> <button book a reservation. toggle open> </button> </div> <div> <select id=1 type> <option reservations true> Dine in </option> <option pickup> Pickup </option> <option delivery> Delivery </option> <option events> Events </option> <option wineries> Wineries </option> <option all> Everything </option> </select> <div id=2> <p> Celebrating and supporting leading women shaking up the industry. </p> <span> Explore now </span> </div> </div> </div> </html>\\ 
\textasciigrave\textasciigrave\textasciigrave\\[1ex] 
Based on the HTML webpage above, try to complete the following task:\\ 
Task: Check for pickup restaurant available in Boston, NY on March 18, 5pm with just one guest\\ 
Previous actions:\\ None\\ 
What should be the next action? Please select from the following choices (If the correct action is not in the page above, please select A. 'None of the above'):\\[1ex] 
A. None of the above\\ 
B. <button id=0 book a reservation. toggle open> <span> Book a\\ 
C. <select id=1 type> <option reservations true> Dine in </option> <option\\ 
D. <div id=2> <p> Celebrating and supporting leading women shaking up\\[1ex] 
Assistant:\\ 
Answer: C.\\ Action: SELECT\\ Value: Pickup \\
\bottomrule
\end{tabularx}
\caption{System Message for GPT-3.5 Planning}
\label{table:gpt_system_message}
\end{table*}

%% file: prompts/reflexion.tex
\begin{table*}[t]
\small
\centering
\begin{tabularx}{\textwidth}{p{1.25cm}X}
    \toprule
    \multicolumn{1}{c}{\textbf{Role}} & \multicolumn{1}{c}{\textbf{Content}} \\
    \midrule
    system & You are an advanced reasoning agent who specializes in analyzing conversational web navigation. You will be presented with a conversation between users and assistants involving a webpage's HTML content, a user's query, and an AI assistant's response. Your objective is to provide a concise and clear one-sentence rationale that explains how the assistant's response is derived from the HTML content in relation to the user's specific query. \\ \midrule
    user & 
    \#\#\# Conversation\newline Webpage: <select id=1 type> <option reservations true> Dine in </option> <option pickup> Pickup </option> <option delivery> Delivery </option> <option events> Events </option> <option wineries> Wineries </option> <option all> Everything </option> </select>\newline \newline User: Based on the HTML webpage above, try to complete the following task:\newline Task: Check for pickup restaurant available in Boston, NY on March 18, 5pm with just one guest.\newline Previous actions:\newline None\newline What should be the next action?\newline \newline Assistant: Action: SELECT\newline Value: Pickup\newline \newline \#\#\# Rationale \\
    & \\
    assistant & The assistant chose to select "Pickup" directly corresponds to the user's request to check for a pickup restaurant. This action is informed by the HTML structure of the webpage, which contains a dropdown menu (<select id=1 type>) with various options including "Pickup" (<option pickup>). \\ \midrule
    user & \#\#\# Conversation\newline Webpage: <button id=2 selected pick-up date 03/19/2023> <span> <span> 19 </span> <div> <span> Mar </span> <span> 2023 </span> </div> </span> </button>\newline \newline User: Based on the HTML webpage above, try to complete the following task:\newline Task: Find a mini van at Brooklyn City from April 5th to April 8th for a 22 year old renter.\newline Previous actions:\newline [searchbox]  Pick-up \& Return Location (ZIP, City or Airport) (... -> TYPE: Brooklyn\newline [option]  Brooklyn, NY, US Select -> CLICK\newline What should be the next action?\newline \newline Assistant: Action: CLICK\newline \newline \#\#\# Rationale \\
    & \\
    assistant & The assistant's response to perform a "CLICK" action is given the user's progress in the task. The user has already selected a location (Brooklyn) for picking up a minivan. The next logical step in the process would be to click the button to select the pick-up date. \\

    \bottomrule
\end{tabularx}
\caption{Prompts for Memory Refinement}
\label{table:reflection_prompt}
\end{table*}

%% file: prompts/case_study.tex
\begin{table*}[t]
\small
\centering
\begin{tabularx}{\textwidth}{p{4cm}X}
    \toprule
    \multicolumn{1}{c}{\textbf{Previous Instruction}} & \multicolumn{1}{c}{\textbf{Action Representations}} \\
    \midrule
    Search for new laptops. &  [combobox]  Search for anything -> TYPE: laptop \newline
    [button]  Search -> CLICK
    \\ \midrule
    Set price from \$400 to \$500. & [input]   -> CLICK \newline
    [textbox]  Minimum Value in \$ -> TYPE: 400 \newline
    [textbox]  Maximum Value in \$ -> TYPE: 500  \newline
    [button]  Submit price range -> CLICK 
    \\ \midrule
    Search for free shipping. & [input]   -> CLICK
    \\ \bottomrule
\end{tabularx}
\caption{Conversation History}
\label{table:case_conversation_history}
\end{table*}

\begin{table*}[t]
\small
\centering
\begin{tabularx}{\textwidth}{X}
\toprule
\textbf{History Trajectory 2} \\
\#\#\# Human: \textasciigrave\textasciigrave\textasciigrave\\

(html (body (header banner (input id=0 submit search ) ) (div main (div group - (ul id=1 (a (img nintendo switch red / blue ) (span \$166.00 ) ) (a (img nintendo switch hac-001 neon yellow! ) (span \$99.99 ) ) $[\ldots]$

\textasciigrave\textasciigrave\textasciigrave\\[1ex]

Based on the HTML webpage above, try to complete the following task:\\
Task: Search for new laptops.\\
What should be the next action? Please select from the following choices (If the correct action is not in the page above, please select A. 'None of the above'):\\\\

A. None of the above \\
B. (input id=0 submit search )\\
C. (ul id=1 (a (img nintendo switch red / blue )\\
D. (div id=2 (button button go to previous slide - )\\
E. (a id=3 eBay Community )\\
F. (a id=4 Accessibility )\\[1ex] 

\#\#\# Assistant: 
B.\\
Action: CLICK\\
\color{blue}{Rationale: The assistant's response to perform a "CLICK" action is derived from the user's task to search for new laptops, and the previous action of typing "laptop" into the search box. Clicking the search button is the next logical step to initiate the search for new laptops based on the HTML structure provided.} \\[1ex]

\midrule
\textbf{History Trajectory 1} \\
\#\#\# Human: \textasciigrave\textasciigrave\textasciigrave\\
(html (body (header banner (td (input id=0 combobox text search for anything \_nkw ) (ul listbox (li id=1 (a option vintage clothing recent searches (b vintage clothing ) ) ) ) ) ) (div main (div group - (a (img id=2 diamond stripe comforter set by ) (span \$34.99 ) ) ) $[\ldots]$

\textasciigrave\textasciigrave\textasciigrave\\[1ex]

Based on the HTML webpage above, try to complete the following task:\\
Task: Search for new laptops.\\
What should be the next action? Please select from the following choices (If the correct action is not in the page above, please select A. 'None of the above'):\\\\

A. None of the above \\
B. (input id=0 combobox text search for anything \_nkw ) \\
C. (li id=1 (a option vintage clothing recent searches (b vintage \\
D. (img id=2 diamond stripe comforter set by ) \\ 
E. (a id=3 (img down alternative forget me not ) (span \\
F. (li id=4 (a Contact Us ) ) \\[1ex] 

\#\#\# Assistant: 
B. \\
Action: TYPE \\
Value: laptop \\
\color{blue}{Rationale: The assistant's response to type "laptop" is derived from the user's task to search for new laptops, and the HTML structure of the webpage, which contains an input element with the id "0" and the type "combobox text" for searching. Therefore, the next action is to type "laptop" into the search input field.} \\[1ex] 

\midrule
\textbf{History Trajectory 7} \\
\#\#\# Human: \textasciigrave\textasciigrave\textasciigrave\\

$[\ldots]$ (li lh\_fs (div (input id=1 checkbox free shipping on ) (span Free Shipping ) ) ) ) (div main (div id=2 (div (img shop on ebay ) (div (a (span heading Shop on eBay ) (span Opens in a new window or tab ) ) (span Brand New ) $[\ldots]$

\textasciigrave\textasciigrave\textasciigrave\\[1ex]

Based on the HTML webpage above, try to complete the following task:\\
Task: Search for free shipping.\\
What should be the next action? Please select from the following choices (If the correct action is not in the page above, please select A. 'None of the above'):\\\\

A. None of the above\\
B. (li id=0 (div button (h3 Buying Format ) ) (ul\\
C. (input id=1 checkbox free shipping on )\\
D. (div id=2 (div (img shop on ebay ) (div (a\\
E. (a id=3 (div 64 GB (span - apply Shop by\\
F. (a id=4 Your Privacy Choices )\\[1ex] 

\#\#\# Assistant: 
C.\\
Action: CLICK\\
\color{blue}{Rationale: The assistant's response to click is derived from the user's task to search for free shipping, which is represented by the checkbox with the label "Free Shipping" in the HTML content. Clicking on this checkbox would allow the user to search for products with free shipping.} \\[1ex]

\bottomrule
\end{tabularx}
\caption{History Trajectory for self-MAP}
\label{table:case_history_trajectory}
\end{table*}

\begin{table*}[t]
\small
\centering
\begin{tabularx}{\textwidth}{X}
\toprule
\textbf{MindAct + CAR} \\
\#\#\# Human: \textasciigrave\textasciigrave\textasciigrave

(html (body (button id=0 button Open (svg id=1 (desc Expand Svg ) ) ) (header banner (tr (input id=2 combobox text search for anything \_nkw ) (input id=3 submit search ) ) ) (div main (div (a Video Game Consoles ) (a id=4 see all video game consoles See all ) ) ) ) )

\textasciigrave\textasciigrave\textasciigrave\\[1ex]

Based on the HTML webpage above, try to complete the following task:\\
Task: \color{blue}{Search for the \textquotesingle xbox series x console\textquotesingle 
 \space with free shipping and a price range from \$400 to \$500.}\\
Previous actions:\\
None \\
What should be the next action? Please select from the following choices (If the correct action is not in the page above, please select A. 'None of the above'):\\\\

A. None of the above\\
B. (button id=0 button Open (svg id=1 (desc Expand Svg )\\
C. (svg id=1 (desc Expand Svg ) )\\
D. (input id=2 combobox text search for anything \_nkw )\\
E. (input id=3 submit search ) \\
F. (a id=4 see all video game consoles See all )\\[1ex]
\#\#\# Assistant:
E.\\
Action: CLICK\\[1ex]
\midrule
\textbf{Synapse} \\
\#\#\# Human: \textasciigrave\textasciigrave\textasciigrave

(html (body (header banner (tr (input id=0 combobox text search for anything \_nkw ) (input id=1 submit search ) ) ) (div main (li (a id=2 Electronics ) (button Expand: Electronics ) ) ) ) )

\textasciigrave\textasciigrave\textasciigrave\\[1ex]

Based on the HTML webpage above, try to complete the following task:\\
Task: Search \textquotesingle xbox series x console\textquotesingle.\\
Previous actions: \\
None \\
What should be the next action? Please select from the following choices (If the correct action is not in the page above, please select A. 'None of the above'):\\\\

A. None of the above\\
B. (input id=0 combobox text search for anything \_nkw )\\
C. (input id=1 submit search )\\
D. (a id=2 Electronics )\\[1ex]
\#\#\# Assistant:
C.\\
Action: CLICK\\[1ex]
\midrule
\textbf{Self-MAP} \\
\#\#\# Human: \textasciigrave\textasciigrave\textasciigrave

(html (body (header banner (tr (input id=0 combobox text search for anything \_nkw ) (input id=1 submit search ) ) ) (div main (li (a id=2 Electronics ) (button Expand: Electronics ) ) ) ) )

\textasciigrave\textasciigrave\textasciigrave\\[1ex] 

Based on the HTML webpage above, try to complete the following task:\\
Task: Search \textquotesingle xbox series x console\textquotesingle.\\
Previous actions:\\
None \\
What should be the next action? Please select the element to interact with, and the action to perform along with the value to type in or select. If the task cannot be completed, output None.\\\\
\#\#\# Assistant:
Element: (input id=0 combobox text search for anything \_nkw ) \\
Action: TYPE\\
Value: xbox series x console \\[1ex]
\bottomrule
\end{tabularx}
\caption{Final Evaluation Steps}
\label{table:case_last_conversation}
\end{table*}